\documentclass[letterpaper]{article} 
\usepackage{aaai2026}  
\usepackage{times}  
\usepackage{helvet}  
\usepackage{courier}  
\usepackage{amsmath} 
\usepackage[hyphens]{url}  
\usepackage{graphicx} 
\usepackage{natbib}  
\usepackage{caption} 
\usepackage{algorithm}
\usepackage{algorithmic}

\usepackage{setspace}
\usepackage[font=footnotesize]{subfig}
\usepackage{overpic}
\usepackage{makecell}
\usepackage{multirow}
\usepackage[table]{xcolor}
\definecolor{mygreen}{HTML}{E6F8E0}
\definecolor{myblue}{HTML}{CFDFF5}
\definecolor{myred}{HTML}{FFCCCC}
\usepackage{pifont}
\usepackage{booktabs}
\usepackage{amssymb}
\usepackage{graphicx}  
\usepackage{anyfontsize}

\usepackage{siunitx}
\usepackage{newfloat}
\usepackage{listings}
\DeclareCaptionStyle{ruled}{labelfont=normalfont,labelsep=colon,strut=off} 
\floatstyle{ruled}
\newfloat{listing}{tb}{lst}{}
\floatname{listing}{Listing}
\title{Audio-Assisted Face Video Restoration\\with Temporal and Identity Complementary Learning}
\author{
    Yuqin Cao\textsuperscript{\rm 1}, 
    Yixuan Gao\textsuperscript{\rm 1}, 
    Wei Sun\textsuperscript{\rm 2}, 
    Xiaohong Liu\textsuperscript{\rm 1}, 
    Yulun Zhang\textsuperscript{\rm 1}, 
    Xiongkuo Min\textsuperscript{\rm 1}\textsuperscript{\rm *}
}
\affiliations{
    \textsuperscript{\rm 1}Shanghai Jiao Tong University\\
    \textsuperscript{\rm 2}East China Normal University\\
    minxiongkuo@sjtu.edu.cn
}

\usepackage{bibentry}

\begin{document}
\maketitle
\begin{abstract}
Face videos accompanied by audio have become integral to our daily lives, while they often suffer from complex degradations. Most face video restoration methods neglect the intrinsic correlations between the visual and audio features, especially in mouth regions. A few audio-aided face video restoration methods have been proposed, but they only focus on compression artifact removal. In this paper, we propose a \textbf{G}eneral \textbf{A}udio-assisted face \textbf{V}ideo restoration \textbf{N}etwork (GAVN) to address various types of streaming video distortions via identity and temporal complementary learning. Specifically, GAVN first captures inter-frame temporal features in the low-resolution space to restore frames coarsely and save computational cost. Then, GAVN extracts intra-frame identity features in the high-resolution space with the assistance of audio signals and face landmarks to restore more facial details. Finally, the reconstruction module integrates temporal features and identity features to generate high-quality face videos. Experimental results demonstrate that GAVN outperforms the existing state-of-the-art methods on face video compression artifact removal, deblurring, and super-resolution. Codes will be released upon publication.
\end{abstract}

\section{Introduction}
\label{sec:intro}
Audio-visual (A/V) streaming services continue to grow in popularity and are rapidly becoming a crucial source of information in our daily lives. In A/V content, the speaker's voice often garners the most attention, leading viewers to instinctively focus on the speaker's face. However, in real-world scenarios, face videos often suffer from complex degradations. Face video restoration methods aim to restore degraded face videos to the authentic, high-quality and reliable ones. This not only improves users' quality of experience (QoE) but also contributes to advancements in video compression technologies and supports visual tasks such as face recognition \cite{kong2021real,lau2021atfacegan,zhang2011close}, privacy protection \cite{yu2016iprivacy}, and autonomous driving \cite{chen2015deepdriving}.

\begin{figure}[!tb]
\captionsetup[subfigure]{justification=centering}
\centering
  \includegraphics[width=\linewidth]{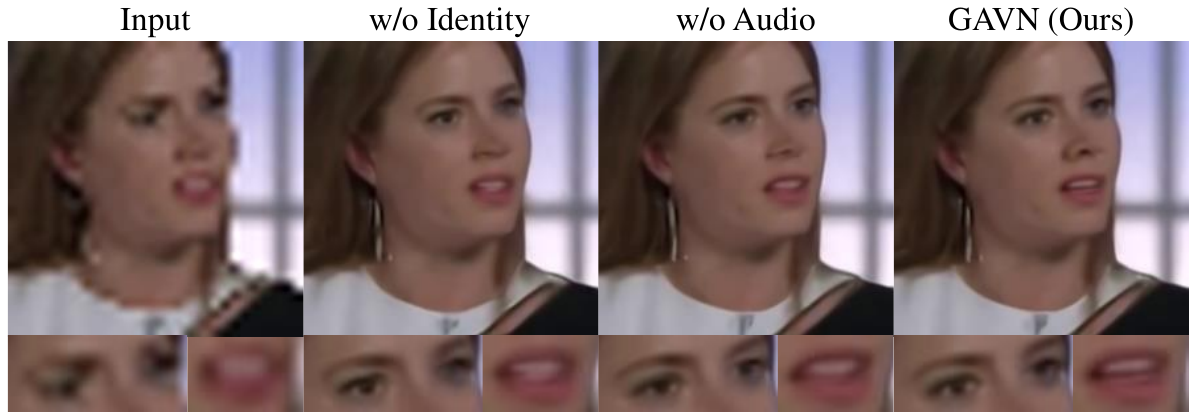}
  \caption{Restoration results of our proposed GAVN with and without identity features and audio signals. The frames reconstructed without using identity features and audio signals appear blurry in the left eye and tongue, respectively.}
\vspace{-0.5cm}
\label{fig:first}
\end{figure}
In the literature, most face restoration studies \cite{wang2019edvr,chan2022basicvsr++,wang2023dr2,yang2021gan} only consider the image/video mode, while ignoring the crucial impact of audio. The motivation of audio-aided face video restoration is the high correlation between visual and audio features,  particularly the synchronization between audio and lip movements in face videos. Physiologically, sound production involves shaping it with facial muscles, particularly those in the lips that control airflow. Therefore, audio can significantly improve face video restoration quality whilst consuming minimal storage space. Moreover, the restored face video should strive to maintain the original appearance under various types of face actions. This is crucial since the face is highly structured and frequently used for personal information identification. It inspires us to leverage facial identity features to aid in video restoration.

Some researchers \cite{zhang2020davd,guo2020deep,zhang2022multi} have explored audio-assisted compressed face video restoration. However, there are still some issues that need to be addressed. Firstly, the existing audio-assisted face video restoration methods are exclusively designed to restore compressed face videos, neglecting the potential of utilizing audio for the restoration of other common distortion types. We propose a novel network, called \textbf{G}eneral \textbf{A}udio-assisted face \textbf{V}ideo Restoration \textbf{N}etwork (GAVN) for the face video compression artifact removal, deblurring, and super-resolution. Secondly, in previous video restoration methods \cite{wang2019edvr,zhou2022revisiting,li2021arvo,guo2022differentiable}, input frames are often downsampled to align and capture features in the low-resolution space. Though this approach significantly reduces the computational cost, it results in the loss of facial detail features. In contrast, our GAVN extracts temporal features from multiple consecutive frames in the low-resolution space and captures identity features with the assistance of audio from individual concurrent frames in the high-resolution space. This allows us to preserve facial details whilst achieving computational efficiency. As illustrated in Fig. \ref{fig:first}, both identity features and audio signals contribute to the restoration of facial details. Thirdly, incorporating face landmark features can further keep the identification information of the restored face \cite{chen2018fsrnet,li2018learning,bulat2018super}. Conventional face landmark detection methods are trained on high-quality face images and encounter difficulties in accurately detecting landmarks from low-quality face images. Therefore, we retrain the landmark detection model to precisely extract landmark features from low-quality face frames with the assistance of audio.

In this paper, we make three contributions to the face video restoration field.
\begin{figure*}[!tb]
\centering
   \includegraphics[width=0.9\linewidth]{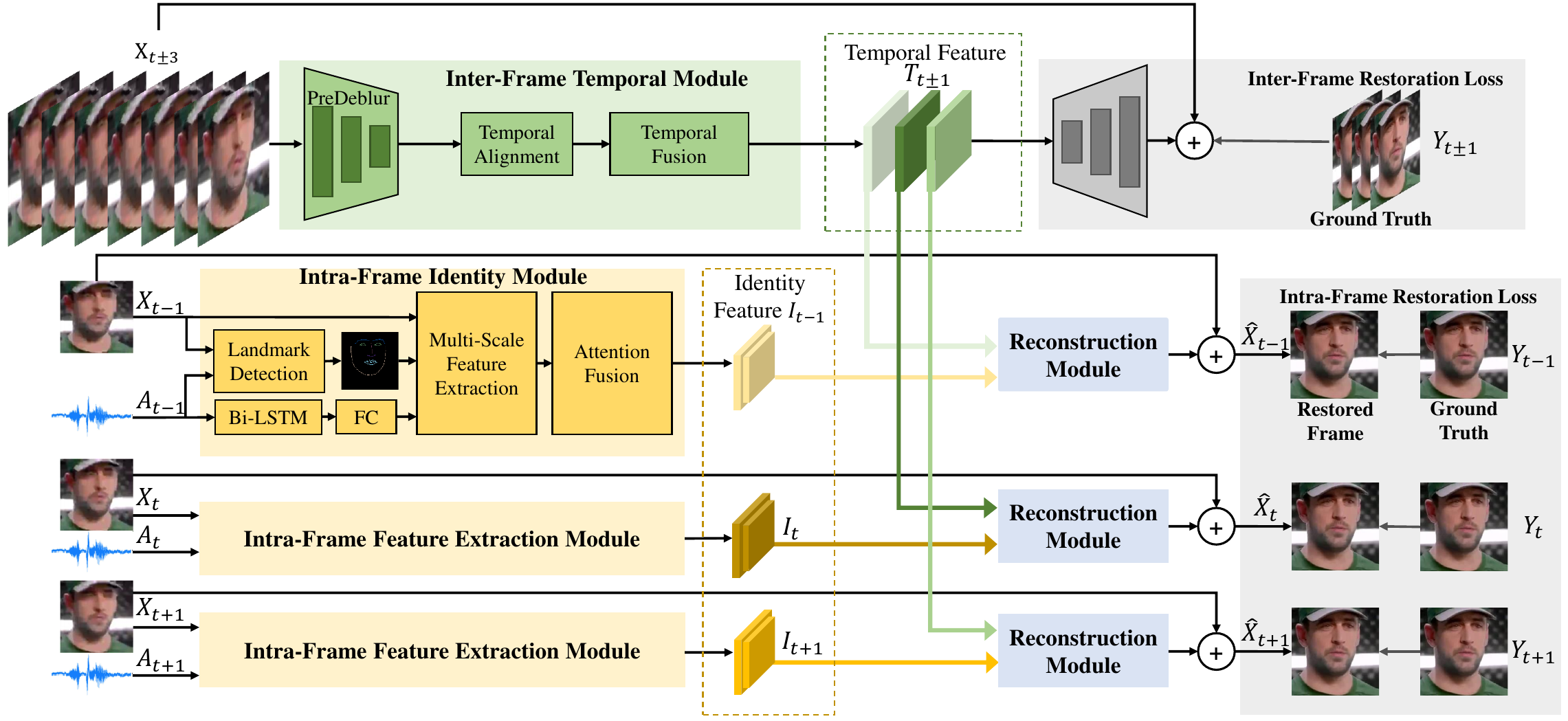}
   \caption{The framework of the proposed GAVN, which consists of three modules: (a) Inter-Frame Temporal Module (Sec. \ref{sec:inter-frame}): extract temporal features from the multiple consecutive frames. (b) Intra-Frame Identity Module (Sec. \ref{sec:intra-frame}): extract identity features with the assistance of audio and landmark features from the single frame. (c) Reconstruction Module (Sec. \ref{sec:reconstruction}): integrate temporal features and identity features to restore high-quality frames.}
\vspace{-0.5cm}
\label{fig:model}
\end{figure*}
\begin{itemize}
    \item We propose GAVN for the face video compression artifact removal, deblurring, and super-resolution.
    \item Our GAVN integrates inter-frame temporal features and intra-frame identity features to restore face videos. With the assistance of audio and identity features, our GAVN outperforms the state-of-the-art (SOTA) video restoration methods.
    \item We conduct experimental analysis to systematically evaluate GAVN, including two scenarios: one involving various speakers with different identities and another involving a specific known speaker.
\end{itemize}

\section{Related Work}
\label{sec:Related Work}

\subsection{Deep Video Restoration}
Most video restoration methods are based on convolutional neural networks (CNNs), which can be divided into two categories: sliding window-based and recurrent-based methods. Sliding window-based methods \cite{wang2019edvr,isobe2020video,li2021arvo,li2020mucan} typically take a small segment of video frames as input and predict the center frame. Recurrent-based methods \cite{haris2019recurrent,sajjadi2018frame,chan2021basicvsr,chan2022basicvsr++} mainly use previously reconstructed high-quality frames or features for subsequent frame reconstruction. Some researcher utilize the recurrent structure to handle input of various lengths and capture long-term temporal information. For instance, Sajjadi~\textit{et al.}~\cite{sajjadi2018frame} proposed a recurrent approach that integrates the previously estimated high-quality frame and the current low-quality frame to predict the current frame. Chan~\textit{et al.}~\cite{chan2021basicvsr} further utilized the bidirectional recurrent network and expanded it to grid propagation in \cite{chan2022basicvsr++}.

Encouraged by the above research, our GAVN incorporates a recurrent structure to extract inter-frame temporal features, thereby mitigating quality fluctuations across frames. Different from the above methods, GAVN further leverages intra-frame identity features and audio features of the current frames to enhance facial details. Their combination outperforms the performance achieved by individual temporal features.

\subsection{Temporal Alignment and Fusion} 
Due to the motions of the camera and object, the adjacent frames are often misaligned. Currently, many researchers \cite{wang2019edvr,tian2020tdan,luo2021ebsr} utilize deformable convolutions to align the neighboring frames to the reference frames. For example, Tian~\textit{et al.}~\cite{tian2020tdan} proposed TDAN that utilizes deformable convolution to align the reference frame and each supporting frame at the feature level without computing optical flow. Wang~\textit{et al.}~\cite{wang2019edvr} proposed the PCD module which extends TDAN to multi-scale alignment and performs alignment in a coarse-to-fine manner. In this paper, we take a step further by employing deformable convolution to align adjacent frames and skip frames, both forward and backward in time. This allows us to aggregate temporal features from different spatio-temporal locations and directions to obtain more comprehensive and abundant temporal features.

\section{Methodology}
\subsection{Overview}
The overall framework of the proposed GAVN is illustrated in Fig. \ref{fig:model}. GAVN is a general architecture suitable for various face video restoration tasks, including face video compression artifact removal, deblurring, and super-resolution. GAVN takes $2N+5$ consecutive low-quality frames $\boldsymbol{X}_{t\pm (N+2)}=\{X_{t-N-2},...,X_{t+N+2}\}$ and the corresponding audio segments $\boldsymbol{A}_{t\pm m}$ as inputs to predict the high-quality frames $\hat{\boldsymbol{X}}_{t\pm N}$, which closely resemble the ground truth frames $\boldsymbol{Y}_{t\pm N}$. For the super-resolution task, low-quality frames are first upsampled to the same resolution as the ground truth frames through Bicubic interpolation.

GAVN consists of three modules: the inter-frame temporal module, the intra-frame identity module, and the reconstruction module. GAVN first utilizes the inter-frame temporal module to extract temporal features from low-quality frames $\boldsymbol{X}_{t\pm (N+2)}$. It utilizes deformable convolutions for frame alignment forward and backward in time. The inter-frame temporal module can be formulated as:
\begin{equation}
    \boldsymbol{T}_{t\pm N} = \mathrm{G}_{\mathrm{inter\mbox{-}frame}}(\boldsymbol{X}_{t\pm (N+2)}),
\end{equation}
where $\boldsymbol{T}_{t\pm N}$ denotes as the temporal features of video frames $\boldsymbol{X}_{t\pm N}$ and $\mathrm{G}_{\mathrm{inter\mbox{-}frame}}$ is the inter-frame temporal module. Since the audio is highly correlated with the movement of the mouth regions, the intra-frame identity module utilizes the current frame and the corresponding audio segments to extract identity features, that is:
\begin{equation}
    \boldsymbol{I}_{t\pm N} = \mathrm{G}_{\mathrm{intra\mbox{-}frame}}(\boldsymbol{X}_{t\pm N}, \boldsymbol{A}_{t\pm m}),
\end{equation}
where $\boldsymbol{I}_{t\pm N}$ denotes as the identity features of video frames $\boldsymbol{X}_{t\pm N}$ and $\mathrm{G}_{\mathrm{intra\mbox{-}frame}}$ is the intra-frame identity module. The audio also assists in detecting the face landmark, which helps the intra-frame identity module to extract identity features more accurately. Lastly, GAVN combines temporal features and identity features to predict the high-quality frames $\hat{\boldsymbol{X}}_{t\pm N}$:
\begin{equation}
    \hat{\boldsymbol{X}}_{t\pm N} = \mathrm{R}(\boldsymbol{T}_{t\pm N}, \boldsymbol{I}_{t\pm N}) + \boldsymbol{X}_{t\pm N},
\end{equation}
where $R$ is the reconstruction module. GAVN can simultaneously restore multiple high-quality face video frames $\hat{\boldsymbol{X}}_{t\pm N}$. During the model optimization process, GAVN first utilizes inter-frame reconstruction loss to optimize the inter-frame temporal module. It then employs intra-frame reconstruction loss to optimize the entire model, including the inter-frame temporal module, the intra-frame identity module, and the reconstruction module.
\begin{figure*}[!t]
 \centering
   \subfloat[Inter-Frame Temporal Module]{\includegraphics[width=.48\linewidth]{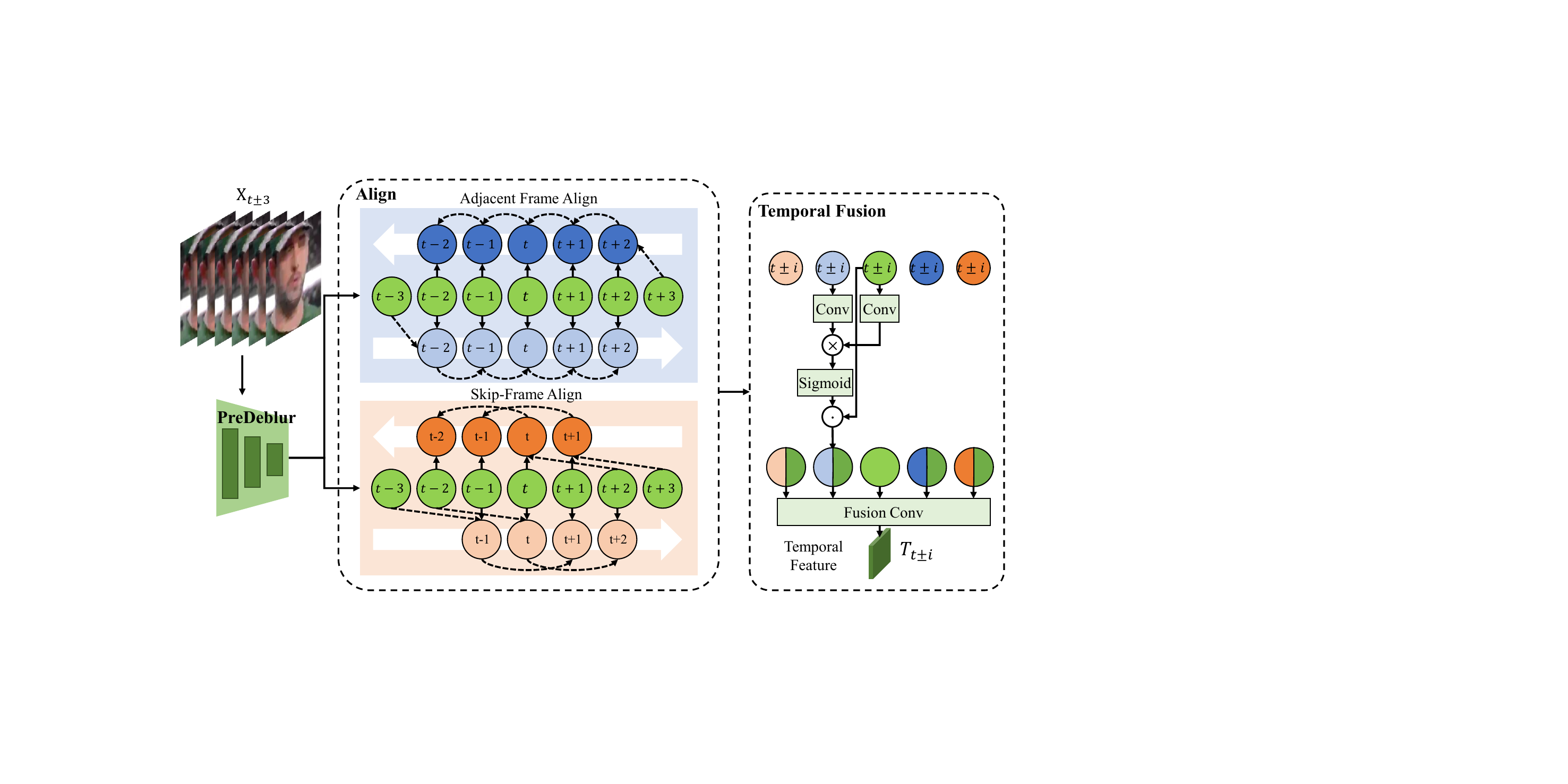}
   \label{fig:inter}}
   \hfil
   \subfloat[Intra-Frame Indentity Module]{\includegraphics[width=.38\linewidth]{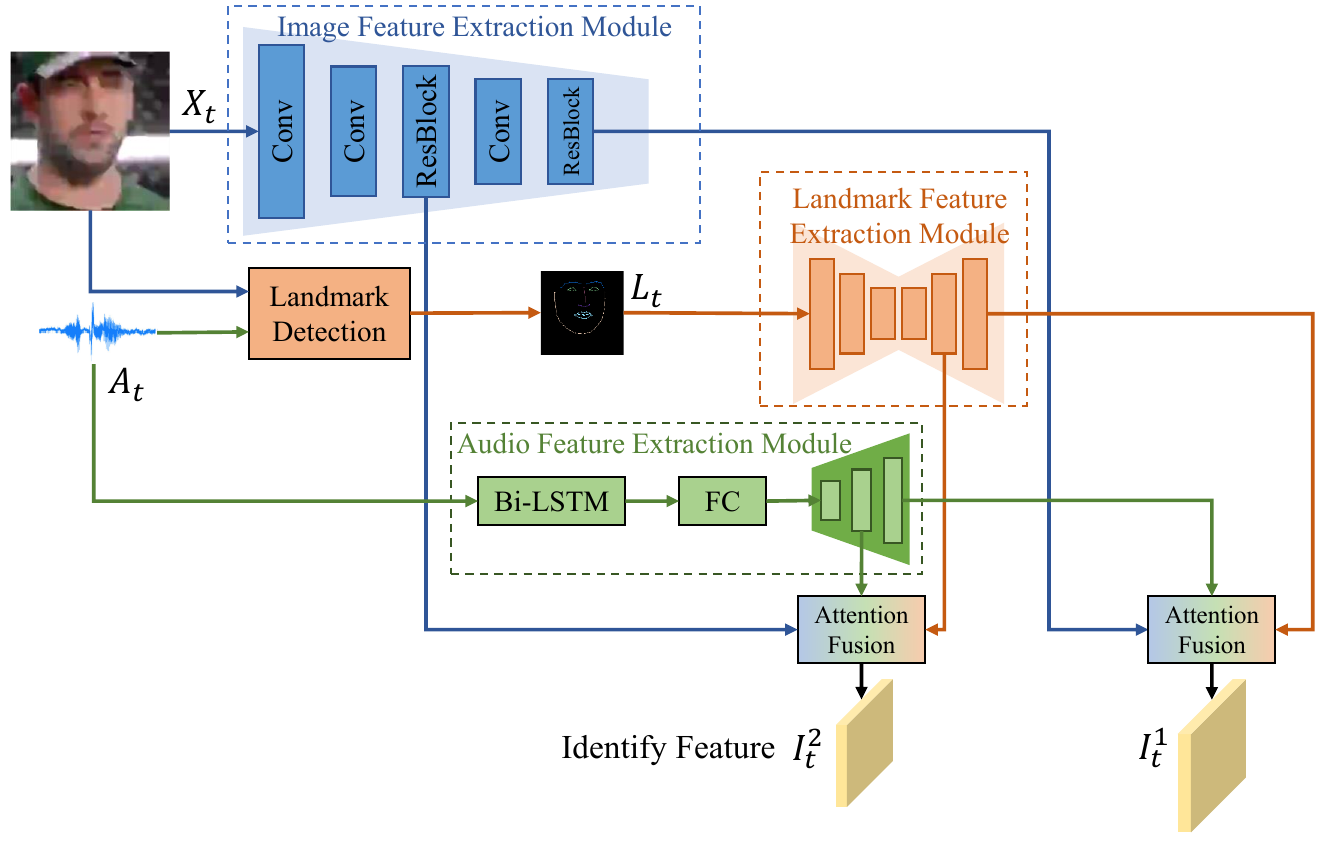}
   \label{fig:intra}}
\vspace{-0.2cm}
\caption{Details of the inter-frame temporal module and the intra-frame identity module. (a) The inter-frame temporal module utilizes deformable convolutions for aligning both adjacent frames and skip frames to extract temporal features. (b) The intra-frame identity module obtains identity features from the single frame aided by audio and landmark features }
\label{fig:intra-inter}
\vspace{-0.4cm}
\end{figure*}
\begin{figure}[!tb]
\captionsetup[subfigure]{justification=centering}
\centering
   \includegraphics[width=\linewidth]{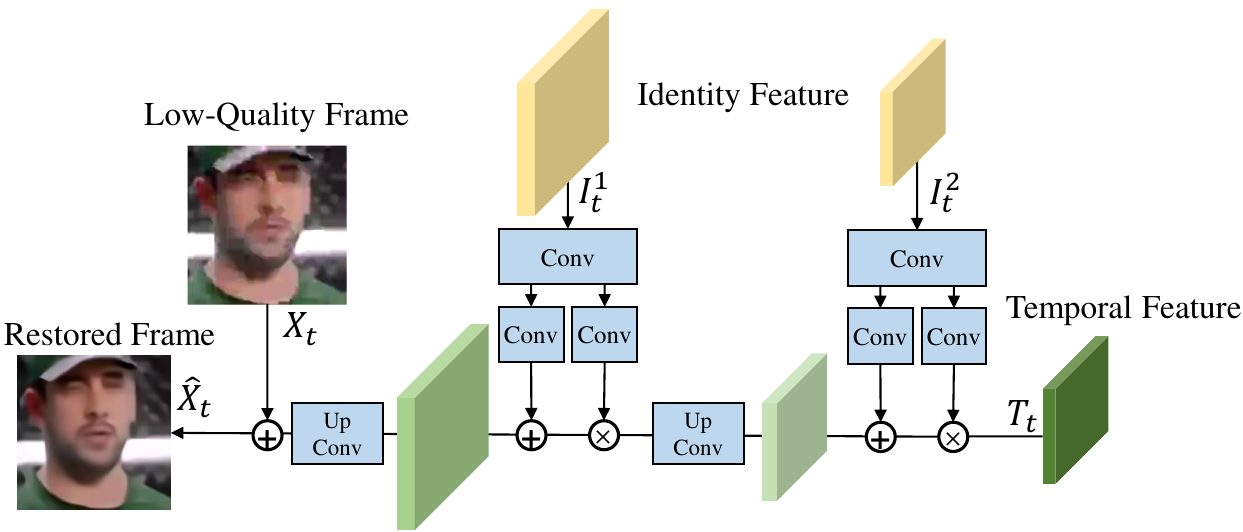}
   \caption{Details of the reconstruction module. It integrates temporal features and identity features to restore high-quality frames.}
\vspace{-0.4cm}
\label{fig:reconstruction}
\end{figure}
\subsection{Inter-Frame Temporal Module}
\label{sec:inter-frame}
In order to fully exploit temporal correlations between adjacent frames in face videos, the inter-frame temporal module extracts temporal features from consecutive low-quality frames in the low-resolution space, thereby saving computational costs. Due to the inevitable movement of the camera and head, there is misalignment between the current frame and its neighboring frames. In order to capture temporal features more accurately, we employ deformable convolutions for both adjacent frame alignment and skip-frame alignment, in both forward and backward time directions. The architecture of the inter-frame temporal module is shown in the left half of Fig. \ref{fig:inter}.

Specifically, the low-quality frames first feed into the predeblur module to downsample input frames with strided convolution layers and obtain $3$-level pyramid features $\boldsymbol{F}_{t\pm N}$ for each frame. Then, we employ deformable convolution to obtain four types of aligned features: forward adjacent, backward adjacent, forward skip-frame, and backward skip-frame aligned features. Here, we take the example of $7$ consecutive frames as inputs:
\begin{equation}
FA_j=\left\{
\begin{array}{rcl}
\mathrm{Align}(F_j, F_{j-1}), & {j=t-2}\\
\mathrm{Align}(F_j, FA_{j-1}), & {j\in[t-1, t+2]}
\end{array} \right.  
\end{equation}
\begin{equation}
BA_j=\left\{
\begin{array}{rcl}
\mathrm{Align}(F_j, F_{j+1}), & {j=t+2}\\
\mathrm{Align}(F_j, BA_{j+1}), & {j\in[t-2, t+1]}
\end{array} \right.  
\end{equation}
\begin{equation}
FS_j=\left\{
\begin{array}{rcl}
\mathrm{Align}(F_j, F_{j-2}), & {j=t-1,t}\\
\mathrm{Align}(F_j, FS_{j-2}), & {j=t+1,t+2}
\end{array} \right.  
\end{equation}
\begin{equation}
BS_j=\left\{
\begin{array}{rcl}
\mathrm{Align}(F_j, F_{j+2}), & {j=t, t+1}\\
\mathrm{Align}(F_j, BS_{j+2}), & {j=t-2, t-1}
\end{array} \right.  
\end{equation}
where $FA_j$, $BA_j$, $FS_j$, and $BS_j$ denote forward adjacent frame aligned features, backward adjacent frame aligned features, forward skip-frame aligned features, and backward skip-frame aligned features of the $j$-th frame, respectively, and $\mathrm{Align}$ represents the alignment operation. We utilize a deformable convolution operation to align the neighboring frame features to the reference frame features from coarse to fine.

Then, temporally fuse the four types of aligned features with the original features to obtain the temporal features. It can pay more attention to the aligned features which are more similar to the original features. We first perform the alignment process directly on the original features to ensure that the aligned features and the original features $F_{t\pm N}$ are of the same size. Then, compute the attention maps between the aligned features and the original features:
\begin{equation}
    AF_j= \mathrm{Align}(F_j, F_j),
\end{equation}
\begin{equation}
    \tilde{Y}_j=AF_j \odot \sigma(\mathrm{Conv}(Y_{j})\times \mathrm{Conv}(AF_j)),
\end{equation}
\begin{equation}
    T_j = \mathrm{Conv}([\tilde{FS}_j , \tilde{FA}_j , \tilde{AF}_j , \tilde{BA}_j , \tilde{BS}_j]),
\end{equation}
where $Y\in\{FA,BA,AF,FS,BS\}$, $\sigma$ is the sigmoid function, $\odot$ is the element-wise multiplication, and $T_j$ is the temporal features of the $j$-th frame.

\subsection{Intra-Frame Identity Module}
\label{sec:intra-frame}
Differing from the inter-frame temporal module, the intra-frame identity module does not include a predeblur module to downsample input frames. It integrates frame features, landmark features, and audio features to obtain identity features from the current single frame in the high-resolution space, as shown in Fig. \ref{fig:intra}. This approach aims to extract facial details more precisely and accurately.

Since most face landmark detection methods are trained on high-resolution face images, it is difficult for them to detect face landmarks from distorted face images. To solve this problem, we pretrain the facial landmark detector network PFLD \cite{guo2019pfld} on the training set. We utilize distorted face frames and the corresponding audio segments as inputs and employ the landmark detection results from the original face frames as ground truth. After pretraining the PFLD network, we can predict relatively accurate face landmarks $\boldsymbol{L}_{t\pm N}$ from distorted face images $\boldsymbol{X}_{t\pm N}$ and the corresponding audio segments $\boldsymbol{A}_{t\pm m}$. Then we utilize the frame feature extraction module, landmark feature extraction module, and audio feature extraction module to extract 2-level frame features, landmark features, and audio features, respectively:
\begin{equation}
    \{F_{f,t}^{1}, F_{f,t}^{2}\} = H_{frame}(X_t),
\end{equation}
\begin{equation}
    \{F_{l,t}^{1}, F_{l,t}^{2}\} = H_{landmark}(L_t),
\end{equation}
\begin{equation}
    \{F_{a,t}^{1}, F_{a,t}^{2}\} = H_{audio}(A_t),
\end{equation}
where $\{F_{f,t}^{1}, F_{f,t}^{2}\}$, $\{F_{l,t}^{1}, F_{l,t}^{2}\}$, and $\{F_{a,t}^{1}, F_{a,t}^{2}\}$ denote $2$-level frame features, landmark features, and audio features of the $t$-th frame, respectively. $H_{frame}$, $H_{landmark}$, and $H_{audio}$ are the frame feature extraction module, landmark feature extraction module, and audio feature extraction module, respectively. 

After obtaining frame, landmark, and audio features, we fuse them to remove distortion and obtain identity features. Inspired by DAVD-Net \cite{zhang2020davd}, we first compute the attention map from the audio and frame features. It indicates which regions of the audio features are more critical for video restoration. The same operation applies to the landmark features. Following this, the audio feature maps and landmark feature maps are element-wise multiplied with their corresponding spatial attention maps, and subsequently combined with the frame feature maps through several convolutional layers, as follows:
\begin{equation}
    \hat{F}^{k}_{l,t} = F^{k}_{l,t}\odot \sigma(Conv([F^{k}_{l,t}, F^{k}_{f,t}])),
\end{equation}
\begin{equation}
    \hat{F}^{k}_{a,t} = F^{k}_{a,t}\odot \sigma(Conv([F^{k}_{a,t}, F^{k}_{f,t}])),
\end{equation}
\begin{equation}
    I^{k}_{t} = Conv([\hat{F}^{k}_{l,t}, F^{k}_{f,t}, \hat{F}^{k}_{a,t}]), k=1,2
\end{equation}
where $k$ denotes the $k$-th level feature. Finally, we obtain $2$-level identity features from the intra-frame identity module.

\begin{table*}[!t]
    \centering
    \caption{Quantitative comparison of GAVN and SOTA restoration methods on the VoxCeleb2 dataset and Obama dataset for three restoration tasks (compression artifact removal, deblur, and super-resolution). The best and second-best performances for each metric are marked in boldface and underlined, respectively.}
    \vspace{-0.3cm}
    \begin{spacing}{1.10}
    \resizebox{\textwidth}{!}{
    \begin{tabular}{c|c|c| cc@{\hspace{0.15cm}}c@{\hspace{0.13cm}}ccc | cc@{\hspace{0.15cm}}c@{\hspace{0.13cm}}ccc}
    \toprule
    \multirow{2}{*}{Task} & \multirow{2}{*}{Method} & \multirow{2}{*}{\makecell[c]{Training \\Frame}} & \multicolumn{6}{c|}{\textbf{VoxCeleb2}} & \multicolumn{6}{c}{\textbf{Obama}}\\ 
    \cline{4-15}
    ~ & ~ & ~ & PSNR$\uparrow$ & SSIM$\uparrow$ & MSSSIM$\uparrow$ & LPIPS$\downarrow$ & Sync$_\text{c}$$\uparrow$ & Sync$_\text{d}$$\downarrow$ & PSNR$\uparrow$ & SSIM$\uparrow$ & MSSSIM$\uparrow$ & LPIPS$\downarrow$ & Sync$_\text{c}$$\uparrow$ & Sync$_\text{d}$$\downarrow$ \\
    \midrule
    \multirow{6}{*}{\makecell[c]{Compression\\Artifact\\Removal}} & DBPN (Haris et al. 2018) & 1 & 28.2960 & 0.8487 & 0.9294 & 0.1749 & 3.6849 & 9.9143 & 27.9026 & 0.8495 & 0.9469 & 0.1397 & 1.2573 & 9.4721\\
    ~ &EDVR \cite{wang2019edvr} & 5 & 28.5213 & 0.8518  & 0.9350 & 0.1686 & 4.3490 & 9.5611 & 28.3826 & 0.8531 & \underline{0.9567} & \underline{0.1163} & 1.4388 & 9.2704 \\
    ~ &BasicVSR++ \cite{chan2022basicvsr++} & 15 & 28.6857 & 0.8513 & 0.9348 & \underline{0.1665} & 4.0944 & 9.7808 & 28.5223 & 0.8469 & 0.9504 & 0.1297 & \underline{1.5376} & \underline{9.1289}\\
    ~ &DAVD-Net \cite{zhang2020davd} & 5 & \underline{28.7269} & \underline{0.8534} & \underline{0.9367} & 0.1694 & \underline{4.6788} & \underline{9.3363} & \underline{28.6645} & \underline{0.8601} & 0.9502 & 0.1330 & 1.3697 & 9.3129 \\
    ~ &VRT \cite{liang2024vrt} & 16 & 28.4900 & 0.8504 & 0.9334 & 0.1703 & 3.9567 & 9.8124 & 28.0689 & 0.8351 & 0.9482 & 0.1349 & 1.2367 & 9.4621\\
    \cline{2-15}
    ~ &\textbf{GAVN}(Ours) & 7 & \textbf{28.9780} & \textbf{0.8622} & \textbf{0.9391} & \textbf{0.1658} & \textbf{4.8463} & \textbf{9.2228}& \textbf{30.0221} & \textbf{0.8898} & \textbf{0.9671} & \textbf{0.1032} & \textbf{1.6664} & \textbf{9.0379}\\
    \midrule
    \midrule
    \multirow{6}{*}{\makecell[c]{Deblur}} & DBPN (Haris et al. 2018) & 1 & 38.0848 & 0.9617 & 0.9903 & 0.0645 & 7.0134 & 7.4523 & 34.5857 & 0.9357 & 0.9872 & 0.0794 & 1.9223 & 8.8230 \\
    ~ &EDVR \cite{wang2019edvr} & 5 & \underline{38.9772} & \underline{0.9698} & \underline{0.9950} & \underline{0.0564} & \underline{7.0534} & 7.4293 & \underline{35.0969} & \underline{0.9446} & \underline{0.9921} & 0.0675 & \underline{1.9918} & \underline{8.7009}\\
    ~ &BasicVSR++ \cite{chan2022basicvsr++} & 15 & 38.5049 & 0.9675 & 0.9945 & 0.0620 & 7.0344 & \underline{7.4287} & 34.7329 & 0.9406 & 0.9915 & 0.0740 & 1.9731 & 8.7216\\
    ~ &DAVD-Net \cite{zhang2020davd} & 5 & 38.4667 & 0.9672 & 0.9943 & 0.0613 & 7.0441 & 7.4355 & 34.9129 & 0.9419 & 0.9913 & \underline{0.0651} & 1.9551 & 8.7518\\
    ~ &VRT \cite{liang2024vrt} & 16 & 38.1804 & 0.9665 & 0.9940 & 0.0629 & 7.0251 & 0.7442 & 34.3923 & 0.9327 & 0.9814 & 0.0824 & 1.9012 & 8.9132\\
    \cline{2-15}
    ~ &\textbf{GAVN}(Ours) & 7 & \textbf{39.3441} & \textbf{0.9716} & \textbf{0.9954} & \textbf{0.0523} & \textbf{7.0603} & \textbf{7.4251}& \textbf{35.2761} & \textbf{0.9461} & \textbf{0.9923} & \textbf{0.0610} & \textbf{2.0543} & \textbf{8.6771} \\
    \midrule
    \midrule
    \multirow{6}{*}{\makecell[c]{Super\\Resolution}} & DBPN (Haris et al. 2018) & 1 & 34.7139 & 0.9394 & 0.9778 & 0.0896 & 6.3998 & 7.8302 & 31.5422 & 0.9176 & 0.9793 & 0.0713 & 1.7924 & 8.9213 \\
    ~ & EDVR \cite{wang2019edvr} & 5 & 35.1852 & 0.9455 & 0.9828 & 0.0725 & \underline{6.6885} & \underline{7.6744}& 32.0530 & 0.9243 & 0.9848 & 0.0575 & 1.8719 & 8.8151  \\
    ~ & BasicVSR++ \cite{chan2022basicvsr++} & 15 & \underline{35.5053} & \underline{0.9496} & \underline{0.9852} & \underline{0.0663} & 6.5442 & 7.7513& \underline{32.9240} & \underline{0.9339} & \underline{0.9868} & \underline{0.0540} & 1.8090 & 8.8738\\
    ~ & DAVD-Net \cite{zhang2020davd} & 5 & 34.5627 & 0.9391 & 0.9797 & 0.0806 & 6.5985 & 7.7462  & 31.9942 & 0.9215 & 0.9837 & 0.0670 & \underline{1.9343} & \underline{8.7629} \\
    ~ & VRT \cite{liang2024vrt} & 16 & 34.9731 & 0.9378 & 0.9765 & 0.0871 & 6.4612 & 7.7419 & 30.2217 & 0.8929 & 0.9782 & 0.0781 & 1.8012 & 8.9552\\
    \cline{2-15}
    ~ & \textbf{GAVN}(Ours) & 7 & \textbf{36.1462} & \textbf{0.9543} & \textbf{0.9859} & \textbf{0.0661} & \textbf{6.7240} & \textbf{7.6504}  & \textbf{33.0620} & \textbf{0.9344} & \textbf{0.9873} & \textbf{0.0488} & \textbf{1.9428} & \textbf{8.7450} \\
    \bottomrule
    \end{tabular}
    }
    \label{tab:overall}
    \vspace{-0.3cm}
    \end{spacing}
\end{table*}
\subsection{Reconstruction Module}
\label{sec:reconstruction}
Temporal features contain richer motion information, whereas identity features capture more spatial details. These features provide different characteristics and compensate for each other. Therefore, we utilize the reconstruction module to integrate both the temporal features and identity features to predict high-quality frames, as shown in Fig. \ref{fig:reconstruction}. The intra-frame identity module captures $2$ level identity features. The $2$-th level identity feature $I_t^2$ has the same size as the temporal feature $T_t$. We first fuse them and upsample to the same size as the $1$-th level identity feature:
\begin{equation}
    \hat{I}_t^2 = \Phi_{\mathrm{UP}} (T_t \times \mathrm{Conv}(I_t^2) + \mathrm{Conv}(I_t^2)),
\end{equation}
where $\Phi_{\mathrm{UP}}$ denotes the upsample residual convolution block, which upsamples the features by a factor of $2$ using the pixel shuffle. The same goes for $1$-th level identity feature:
\begin{equation}
    \hat{I}_t^1 = \Phi_{\mathrm{UP}} (\hat{I}_t^2 \times \mathrm{Conv}(I_t^1) + \mathrm{Conv}(I_t^1)).
\end{equation}
Finally, reduce the channel of the fused features $\hat{I}_t^1$ to $3$ channels, and add to the input low-quality frames $X_t$ to obtain the predicted high-quality frames $\hat{X}_t$.

\begin{figure*}[!tb]
\vspace{-0.1cm}
\captionsetup[subfigure]{justification=centering}
\centering
    \begin{overpic}
   [width=0.88\linewidth]{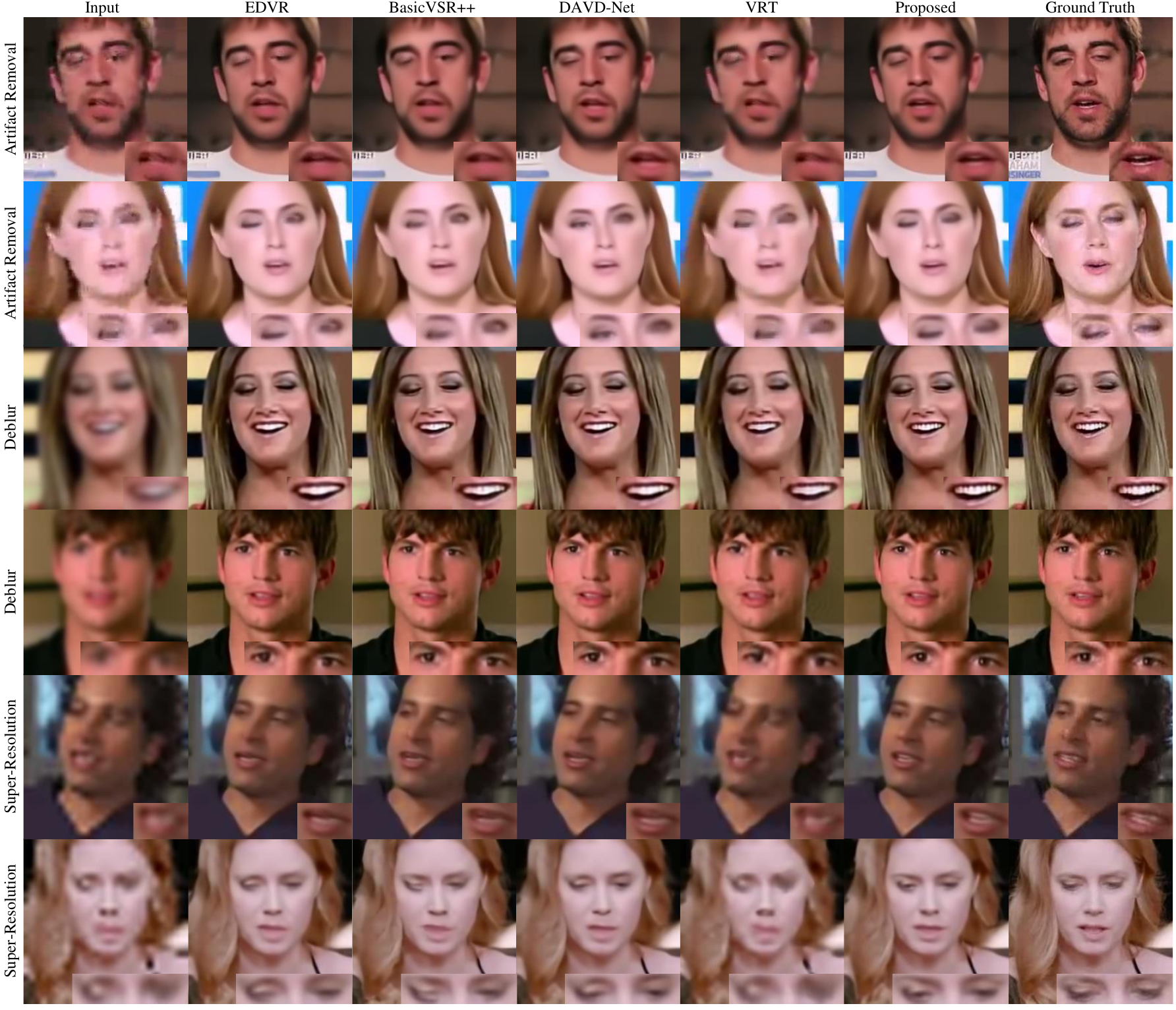}
   \end{overpic}
   \vspace{-0.3cm}
   \caption{Qualitative results on VoxCeleb2 dataset. Distortion types from top to bottom: compression, blur, and low resolution.}
\vspace{-0.3cm}
\label{fig:vox2}
\end{figure*}
\begin{figure*}[!tb]
\captionsetup[subfigure]{justification=centering}
\centering
    \begin{overpic}
  [width=0.88\linewidth]{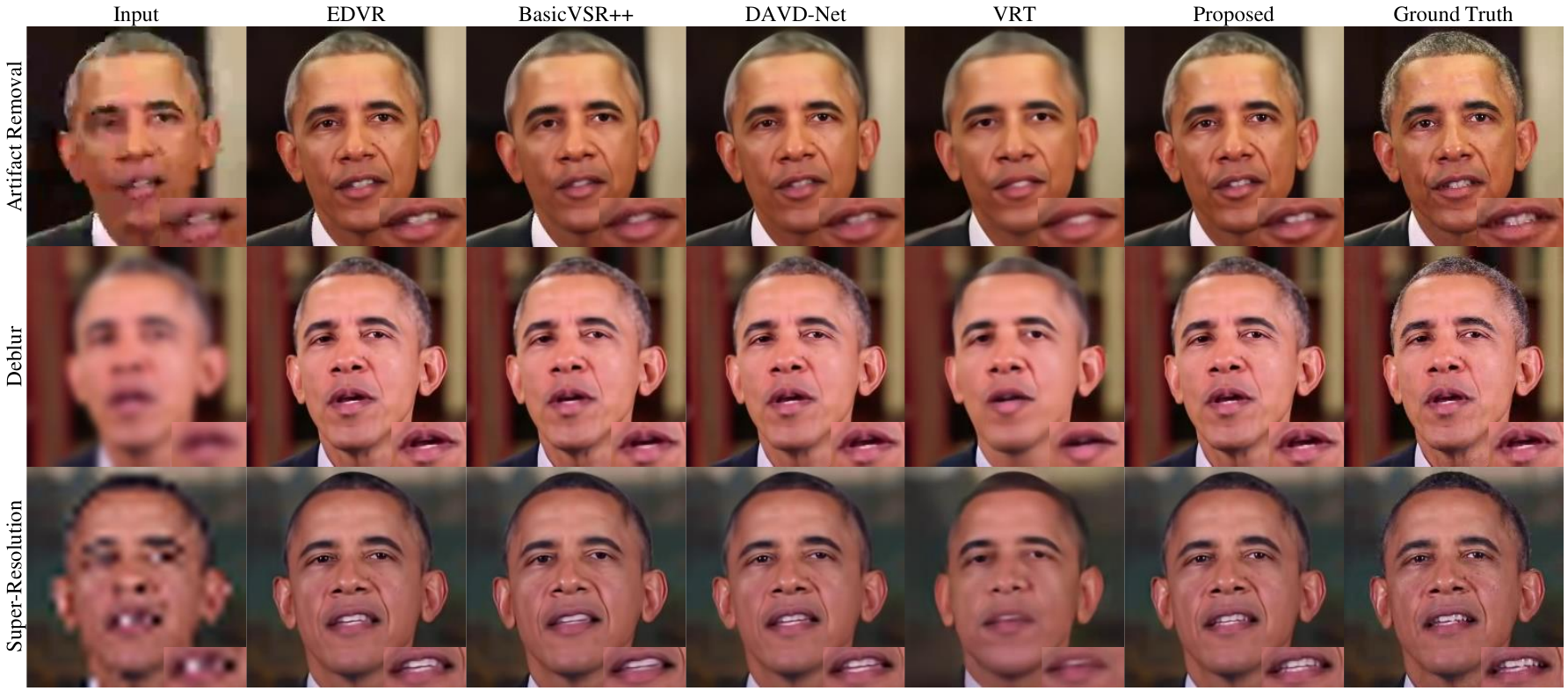}
  \end{overpic}
   \vspace{-0.2cm}
   \caption{Qualitative results on the Obama dataset. Distortion types from top to bottom: compression, blur, and low resolution.}
   \vspace{-0.5cm}
\label{fig:obama}
\end{figure*}
\section{Experiments}
We conduct a comprehensive performance analysis for our GAVN method on two datasets: the VoxCeleb2 dataset \cite{chung2018voxceleb2,nagrani2017voxceleb} and the Obama dataset. The VoxCeleb2 and Obama datasets cover multi-speaker and single-speaker scenarios, respectively, ensuring that GAVN can generalize across diverse scenarios. Specifically, we validate GAVN across three types of face video restoration task, including compression artifact removal, deblurring, and super-resolution, to demonstrate its generalizability. Restored video samples are provided in the supplementary materials.

\subsection{Dataset}
VoxCeleb2 contains over 1 million speeches from $6,112$ celebrities. Due to the limited computing resources, 
we randomly select a training set ($200$ celebrities with $4860$ videos), a validation set ($5$ celebrities with $297$ videos) and a testing set
($20$ celebrities with $894$ videos) with no overlap of speakers. The Obama dataset consists of 180 high-quality weekly address videos from the White House website, ranging from one to six minutes. It is split into $150$ videos for training, $5$ videos for validation, and the rest $25$ videos for testing. We crop and resize the face region of each frame to $224\times 224$ resolution.

To simulate real-world degradation, we apply three types of distortions: compression, blur, and low resolution. Compression videos are generated using FFmpeg with the x264 codec and a CRF of $45$. For blur, we apply a Gaussian filter with a kernel size between $15$ and $25$. For low resolution, we use Bicubic interpolation with random downsampling factors from $2$ to $8$.
\subsection{Evaluation Criteria}
For image quality metrics, we utilize PSNR, SSIM, MSSSIM, and LPIPS. MS-SSIM is an extension of SSIM designed to measure structural similarity at multiple scales. LPIPS utilizes a deep neural network to capture perceptual similarity between images, which can provide a more accurate alignment with human judgments. For lip-sync quality metric, we use SyncNet’s confidence (Sync$_\text{c}$) and SyncNet’s  distance (Sync$_\text{d}$) \cite{chung2017out}. Sync$_\text{c}$ can assess the synchronization quality between the restored face video and audio. Sync$_\text{d}$ measures the distance between the lip and audio representations. Better face video restoration methods should have larger PSNR, SSIM, MSSSIM, Sync$_\text{c}$ values and smaller LPIPS, Sync$_\text{d}$ values.

\subsection{Training Details}
The training procedure of GAVN consists of two steps. In the first step, we only optimize the inter-frame temporal module. We only extract the temporal features from the inter-frame temporal module to predict the restored frames. We utilize the Charbonnier penalty function \cite{lai2017deep} as the loss function. The training epoch is set to $20$ epochs. The learning rate is set to $4\times10^{-4}$. 


In the second step, we also utilize the Charbonnier penalty function as the loss function to optimize the entire model. We implement a warm-up training strategy: we utilize the inter-frame temporal module pretrained in the first step to extract temporal features and only train the intra-frame identity module and the reconstruction module in the first $5$ epochs with the learning rate $lr=4\times10^{-4}$. For the subsequent $15$ epochs, we fine-tune the entire model with the learning rate $lr=2\times10^{-4}$, including the inter-frame temporal module, the intra-frame identity module, and the reconstruction module.

We train the GAVN using the Adam optimizer \cite{kingma2014adam} with $\beta_1 = 0.9$ and $\beta_2 = 0.999$. We implement the proposed GAVN model in PyTorch \cite{ketkar2021introduction} and train it with one NVIDIA A100 GPU.

\begin{table}[!t]
    \centering
    \caption{PSNR comparison with SOTA methods on the VoxCeleb2 dataset under different distortion levels. $\sigma$ represents the CRF scale, Gaussian blur level, and downsampling scale respectively.}
    \vspace{-0.4cm}
    \resizebox{0.5\textwidth}{!}{
    \begin{tabular}{c|c||cccc}
    \toprule
    Distortion & $\sigma$ & EDVR & BasicVSR++  & DAVD-Net  & GAVN\\
    \midrule
    \multirow{3}{*}{Compression} & 35 & 31.5020 & 31.6120 & 31.7197 & \textbf{31.8327} (+0.1130) \\
    ~ & 40 & 30.7648 & 30.8615 & 30.8785 & \textbf{31.0203} (+0.1418)\\
    ~ & 45 & 28.5213 & 28.6857 & 28.7269 & \textbf{28.9780} (+0.2923)\\
     \midrule
    \multirow{3}{*}{Blur} & 17 & 39.2855 & 38.9902 & 38.7835 & \textbf{39.6316} (+0.3461)\\
    ~ & 21 & 38.2728 & 37.8910 & 37.7774 & \textbf{38.7089} (+0.4361)\\
    ~ & 25 & 37.1521 & 36.2650 & 36.5882 & \textbf{37.5581} (+0.4060)\\
     \midrule
    \multirow{3}{*}{Low Resolution} & 2 & 43.1496 & 42.4803 & 42.9954 & \textbf{44.1499} (+1.0003) \\
    ~ & 4 & 36.2302 & 36.5015 & 35.3814 & \textbf{37.1964} (+0.6949)\\
    ~ & 8 & 29.2927 & 29.7503 & 28.6765 & \textbf{30.1159} (+0.3656)\\
    \bottomrule
    \end{tabular}
    }
    \vspace{-0.3cm}
    \label{tab:ablation3}
\end{table}
\begin{table}[!t]
    \centering
    \caption{Quantitative results of ablation studies on the VoxCeleb2 dataset. IF: identity feature; AF: audio feature.}
    \vspace{-0.3cm}
    \begin{spacing}{1.10}
    \resizebox{0.5\textwidth}{!}{    
    \begin{tabular}{c|c|cc@{\hspace{0.15cm}}c@{\hspace{0.13cm}}ccc}
    \toprule
    Task & Method  & PSNR$\uparrow$ & SSIM$\uparrow$ & MSSSIM$\uparrow$ & LPIPS$\downarrow$ & Sync$_\text{c}$$\uparrow$ & Sync$_\text{d}$$\downarrow$ \\
    \midrule
    \multirow{3}{*}{\makecell[c]{Compression\\Artifact\\Removal}} & \emph{w/o} IF& 28.7797 & 0.8547 & 0.9378 & 0.1759 & 4.7170 & 9.3214\\
    ~ & \emph{w/o} AF& 28.9480 & 0.8618 & 0.9382 & 0.1718 & 4.8340 & 9.2663\\
    ~ &\textbf{GAVN}(Ours)& \textbf{28.9780} & \textbf{0.8622} & \textbf{0.9391} & \textbf{0.1658} & \textbf{4.8463} & \textbf{9.2228} \\
    \midrule
    \multirow{3}{*}{\makecell[c]{Deblur}} & \emph{w/o} IF & 38.8352 & 0.9691 & 0.9949 & 0.0579 & 7.0517 & 7.4337 \\
    ~ & \emph{w/o} AF & 39.0181 & 0.9701 & 0.9951 & 0.0540 & 7.0555 & 7.4311 \\
    ~ &\textbf{GAVN}(Ours)& \textbf{39.3441} & \textbf{0.9716} & \textbf{0.9954} & \textbf{0.0523} & \textbf{7.0603} & \textbf{7.4251} \\
    \midrule
    \multirow{3}{*}{\makecell[c]{Super\\Resolution}} & \emph{w/o} IF& 35.7568 & 0.9512 & 0.9821 & 0.0680 & 6.7111 & 7.6538\\
    ~ & \emph{w/o} AF& 35.8427 & 0.9516 & 0.9842 & 0.0678 & 6.7125 & 7.6512\\
    ~ & \textbf{GAVN}(Ours) & \textbf{36.1462} & \textbf{0.9543} & \textbf{0.9850} & \textbf{0.0661} & \textbf{6.7240} & \textbf{7.6504}\\
    \bottomrule
    \end{tabular}
    }
    \vspace{-0.3cm}
    \label{tab:ablation1}
    \end{spacing}
\end{table}
\subsection{Comparison with SOTA Methods}
We compare our GAVN with several SOTA restoration methods: DBPN \cite{haris2018deep}, EDVR \cite{wang2019edvr}, BasicVSR++ \cite{chan2022basicvsr++}, DAVD-Net \cite{zhang2020davd}, and VRT \cite{liang2024vrt}. DBPN is proposed for image super-resolution, EDVR, BasicVSR++, and VRT are the unified framework extensible to various video restoration tasks, and DAVD-Net is designed for the task of audio-aided video compression artifact removal. We utilize the same datasets in our experiments to retrain all compared methods for three restoration tasks for a fair comparison.

The quantitative results on the VoxCeleb2 and Obama datasets are listed in Table \ref{tab:overall}, from which we have several noteworthy observations. Firstly, it is evident that GAVN achieves the best performance across all metrics, particularly in Sync$_\text{c}$ and Sync$_\text{d}$. This indicates GAVN's superior capability to reconstruct the mouth region based on audio information. It can significantly improve synchronization quality between the restored face videos and audio.
Secondly, on the VoxCeleb2 and Obama datasets, GAVN is better than the audio-aided face video restoration method DAVD-Net. This indicates that GAVN can learn better correlations between facial dynamics and speakers' voices and can leverage this knowledge to enhance face restoration. 
Thirdly, compared to the Obama dataset, GAVN has a more significant improvement on the VoxCeleb2 dataset. The larger and more diverse VoxCeleb2 dataset enables GAVN to more effectively improve face video restoration quality.


Qualitative results are presented in Figs. \ref{fig:vox2} and \ref{fig:obama}. GAVN exhibits richer detail recovery compared to other restoration methods, especially in the mouth and eye regions. As shown in the second row of Fig. \ref{fig:vox2}, video restoration methods may mistakenly predict the open and closed states of the eyes. While GAVN can capture finer details and enhance quality of eye regions, providing clearer upper eyelid contour and highlights in the pupils. Moreover, GAVN can better restore the details of the mouth region, including the lips, upper and lower teeth, as well as the gaps between the teeth.

We further conduct experiments at different distortion levels to validate the effectiveness of GAVN across various degree of low-quality videos. The different distortion levels and the corresponding experimental results are listed in Table \ref{tab:ablation3}. It highlights the superiority of GAVN on restoring different quality videos.



\subsection{Ablation Studies}
We conduct ablation studies to validate the importance of intra-frame identity features and audio features in GAVN. The experimental results on the VoxCeleb2 dataset are shown in Table \ref{tab:ablation1}. Removing the intra-frame identity features eliminates the support of landmark and audio features, making GAVN a regular video restoration model. Consequently, its performance is comparable to BasicVSR++. As shown in the third row of each restoration task in Table \ref{tab:ablation1}, removing audio features significantly reduces both the quality of the restored faces and the consistency between speech and lip movements. The qualitative results of the ablation studies can be found in the supplementary materials.

\begin{figure}[!tb]
\captionsetup[subfigure]{justification=centering}
\centering
    \begin{overpic}
  [width=0.95\linewidth]{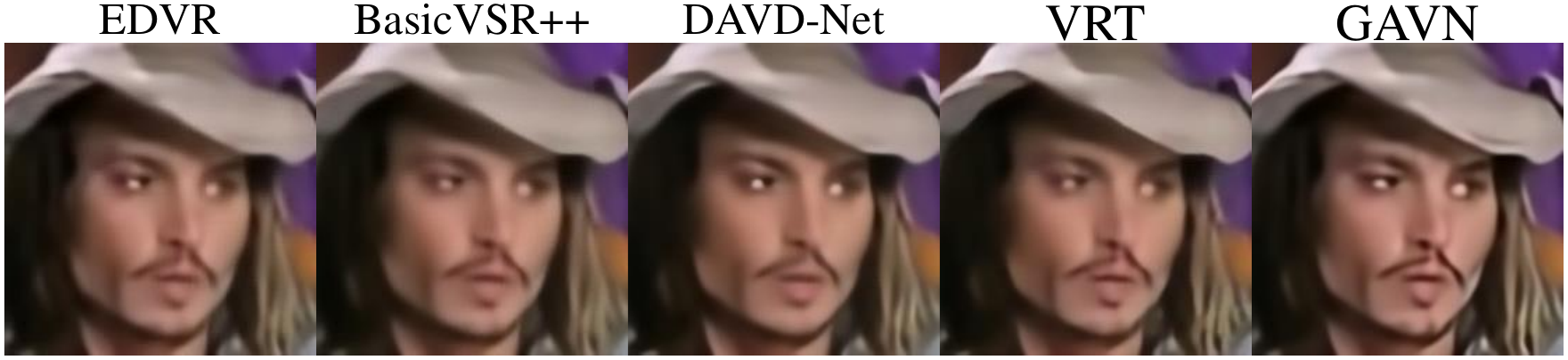}
  \end{overpic}
  \vspace{-0.3cm}
  \caption{Qualitative results on real-world videos.}
  \vspace{-0.3cm}
\label{fig:realworld}
\end{figure}
\begin{table}[!t]
    \centering
    \caption{Quantitative results on real-world videos.}
    \vspace{-0.3cm}
    \resizebox{0.5\textwidth}{!}{
    \begin{tabular}{c|cccccc}
    \toprule
    Metrics & EDVR& BasicVSR++ & DAVD-Net& VRT& GAVN(Ours)\\
    \midrule
    NIQE $\downarrow$ & 6.5407 & 6.2761 & 6.1620 & 6.1762 & \textbf{6.1549}\\
    Sync$_\text{c}$ $\uparrow$ & 1.6590 & 1.6065 & 1.6581 & 1.6326 & \textbf{1.7895}\\
    Sync$_\text{d}$ $\downarrow$ & 7.2050 & 7.3655 & 7.1657 & 7.1825 & \textbf{7.1253}\\
    \bottomrule
    \end{tabular}
    }
    \label{tab:rw}
    \vspace{-0.4cm}
\end{table}
\subsection{Experiments on Real-World Degraded Face Videos}
We collected $10$ real-world degraded face videos from YouTube, featuring diverse genders, skin tones, and hair colors. Since there are no corresponding high-quality undistorted face videos available, we employed NIQE, SyncNet’s confidence score (Sync$_\text{c}$), and average distance (Sync$_\text{d}$) to evaluate the naturalness and audio-visual synchronization of restored face videos. The quantitative and qualitative results on real-world restored face videos are presented in Table \ref{tab:rw} and Fig. \ref{fig:realworld}. It can be observed that our GAVN outperforms other methods, achieving the best performance in the restoration of real-world degraded face videos. This demonstrates that GAVN can better handle complex real-world distortions, making it suitable for a wider range of applications.
\section{Conclusion}
We propose a general audio-assisted face video restoration method GAVN for the face video compression artifact removal, deblurring, and super-resolution. GAVN leverages inter-frame temporal features and intra-frame identity features to restore face videos. Temporal features capture complex inter-frame motion information to restore frames coarsely and then identity features refine more facial details. We conducted experiments in two different scenarios. The results demonstrate that GAVN achieves SOTA performance. The integration of audio and identify features significantly improves reconstruction quality. Moreover, in real-world distortion scenarios, our method outperforms other SOTA methods in restoring face videos.
\bibliography{aaai2026}

\end{document}